\documentclass[a4paper,11pt,twocolumn,twoside]{article}
\usepackage[dvips]{graphicx}
\usepackage{sepln_en}
\usepackage{fullname}
\usepackage{multirow}
\usepackage{url}
\usepackage[utf8]{inputenc}
\usepackage[english]{babel}
\usepackage{multirow}

\usepackage[colorinlistoftodos]{todonotes}

\input epsf

\title{A computational psycholinguistic evaluation of the syntactic abilities of Galician BERT models at the interface of dependency resolution and training time}

\author{\textbf{Iria de-Dios-Flores,} \textbf{Marcos Garcia}\\
Centro Singular de Investigación en Tecnoloxías Intelixentes (CiTIUS)\\ Universidade de Santiago de Compostela\\
\{\texttt{iria.dedios, marcos.garcia.gonzalez\}@usc.gal}
}

\seplntranstitle{Una evaluación psicolingüístico-computacional de las capacidades sintácticas de los modelos BERT para el gallego en la intersección entre la resolución de dependencias y el tiempo de entrenamiento}

\seplnkey{BERT models, Galician, targeted syntactic evaluation, agreement dependencies.}

\seplnabstract{This paper explores the ability of Transformer models to capture  subject-verb and noun-adjective agreement dependencies in Galician. We conduct a series of word prediction experiments in which we manipulate dependency length together with the presence of an attractor noun that acts as a lure. First, we evaluate the overall performance of the existing monolingual and multilingual models for Galician. Secondly, to observe the effects of the training process, we compare the different degrees of achievement of two monolingual BERT models at different training points. We also release their checkpoints and propose an alternative evaluation metric. Our results confirm previous findings by similar works that use the agreement prediction task and provide interesting insights into the number of training steps required by a Transformer model to solve long-distance dependencies.}

\seplnclave{Modelos BERT, Gallego, evaluación sintáctica dirigida, dependencias de concordancia.}

\seplnresumen{Este trabajo analiza la capacidad de los modelos Transformer para capturar las dependencias de concordancia sujeto-verbo y sustantivo-adjetivo en gallego. Llevamos a cabo una serie de experimentos de predicción de palabras manipulando la longitud de la dependencia junto con la presencia de un sustantivo intermedio que actúa como distractor. En primer lugar, evaluamos el rendimiento global de los modelos monolingües y multilingües existentes para el gallego. En segundo lugar, para observar los efectos del proceso de entrenamiento, comparamos los diferentes grados de consecución de dos modelos monolingües BERT en diferentes puntos del entrenamiento. Además, publicamos sus puntos de control y proponemos una métrica de evaluación alternativa. Nuestros resultados confirman los hallazgos anteriores de trabajos similares que utilizan la tarea de predicción de concordancia y proporcionan una visión interesante sobre el número de pasos de entrenamiento que necesita un modelo Transformer para resolver las dependencias de larga distancia.}

\firstpageno{1}

\begin{document}

\setlength\titlebox{21cm} 

\label{firstpage} \maketitle

%

\section{Introduction}

Current language models (LMs) based on deep neural network architectures obtain impressive performance on most NLP tasks, including semantic and syntactic applications \cite{devlin-etal-2019-bertB}. In fact, it has been argued that LSTM and Transformer models may encode syntactic information captured in an unsupervised manner from unlabeled text \cite{lin-etal-2019-open,hewitt-manning-2019-structural}.

To explore the syntactic capabilities of LMs, various studies have probed their grammatical competence by analyzing the resolution of long-distance dependencies using a word prediction task, often known as targeted syntactic evaluation or TSE \cite{linzen-etal-2016-assessing,gulordava-etal-2018-colorless}. Inspired by classical psycholinguistic experiments on human sentence processing, this task consists on comparing the model probabilities for a correct and incorrect alternative in the context of the targeted syntactic phenomena. For instance, given the well-known sentence in psycholinguistic research ``The \textbf{key} to the \underline{cabinets} \textbf{is$|$*are} on the table'', a model that correctly identifies the dependency between the subject (key) and the verb should assign a higher probability to the singular form (is) than to the plural form (are), despite the presence of an intervening plural attractor noun (cabinets).

First studies have shown that recurrent neural networks (RNNs) are able to solve most cases of the agreement prediction task in several languages (though mostly in English) and scenarios \cite{bernardy-lappin-2017-using,gulordava-etal-2018-colorless,kuncoro-etal-2018-lstms}. The interest in the assessment of syntactic abilities of language models grew with the popularization of Transformer architectures \cite{vaswani2017attention}, which learn relations between words using a self-attention mechanism, and seem to perform better than models based on RNNs \cite{devlin-etal-2019-bertB}. In this respect, recent works are putting the focus on the training procedure: \namecite{perez-mayos-etal-2021-much} measure the impact of the amount of training data on syntactic probing, while \namecite{wei-etal-2021-frequency} analyze the influence of word frequency on subject-verb number agreement. 
However, to the best of our knowledge, there are still no studies exploring the models' performance along the training process, i.e., how many training steps do they need to solve long-distance dependencies. This is one of the goals of the present work.

On the evaluation side, it has been argued that instead of comparing the probability of a single correct$|$incorrect pair (as is$|$are in the example above), these experiments should use large lists of pairs representing the same phenomena (e.g., exist$|$exists) to observe the model's \textit{systematicity} (i.e., in how many pairs a model succeeds) and its \textit{likely behaviour} (the probability of generating a correct inflection) \cite{newman-etal-2021-refining}. Nevertheless, this type of evaluation requires large sets of target pairs and, what is more substantial, it assumes a total independence between syntax and semantics ---something which is controversial from a linguistic and psycholinguistic point of view.

Taking the above into account, in this paper we investigate the ability of Transformer models to capture fundamental linguistic operations such as dependency resolution in a less studied language, Galician. Following previous research inspired by both computational modeling and psycholinguistic research, we conduct a series of word prediction experiments using a dataset that targets two types of agreement dependencies (subject-verb dependencies and noun-adjective dependencies) in two experimental conditions (short and long-distance dependencies) while also manipulating the presence of an attractor noun that acts as a lure (e.g. ``O \textbf{neno} que xogaba onte alí coa \underline{nena} é \textbf{alto$|$*alta}'').\footnote{Here \textit{alto} (`tall' in masculine) agrees in its gender features with the correct antecedent \textit{neno} ('boy'), while \textit{alta} (`tall' in feminine) agrees in gender with the structurally irrelevant noun \textit{nena} ('girl'), hence producing an ungrammatical dependency.} First, we evaluate the overall performance of the existing monolingual and multilingual models for Galician. Secondly, in order to observe the effects of the training process, we compare the different degrees of achievement of two monolingual BERT models (which vary on the number of hidden layers, vocabulary size, and initialization) at various training steps.

In addition, we propose an alternative evaluation metric of accuracy, which puts the focus on the probability distance between the correct and the incorrect alternatives for a given semantic plausible word. This metric allows us to explore a model's confidence in producing syntactically and semantically well-formed expressions and to compare models with different vocabulary sizes.

Our contributions are the following: (i) 34 checkpoints of two BERT models for Galician which allow to explore the effects of the training steps on different tasks; (ii) a novel metric for targeted syntactic evaluation which focuses on the probability distribution between correct and incorrect alternatives; (iii) a careful comparison of the models' performance on two agreement dependencies, analyzing the impact of the learning steps, the amount of training data, the model initialization, and the depth of the neural network.

Our results confirm previous findings by similar works using the agreement prediction task and provide interesting insights into the number of training steps required by a Transformer model to solve long-distance dependencies, as they already achieve high performance at early checkpoints when trained on enough data.

The rest of this paper is organized as follows: Section~\ref{sec:related} introduces previous work about targeted syntactic evaluation on neural language models. 
Then, in Section~\ref{sec:models}, we present the main characteristics of the models used for the experiments and the different checkpoints provided by our study. In Section~\ref{sec:method} we describe our methodology, including the rationale behind the evaluation metric proposed here. Finally, the results are presented and discussed in Section~\ref{sec:results}, while Section~\ref{sec:conclusions} draws the conclusions of the work.

\section{Background}
\label{sec:related}
\namecite{linzen-etal-2016-assessing} introduced the \textit{number prediction task} to evaluate the performance of language models on long-distance agreement, and their results suggested that even if LSTM models seem not to generalize syntactic structures, they identify subject-verb agreement dependencies. Inspired by this paper, various studies explored the behaviour of LSTMs models on a variety of languages and syntactic phenomena, arguing that these networks may achieve near-human performance in some agreement experiments \cite{bernardy-lappin-2017-using,gulordava-etal-2018-colorless,kuncoro-etal-2018-lstms} even though they may be alternative explanations (such as surface-based heuristics) that explain the  models' success \cite{kuncoro2018perils,linzen-brian18}. Moving forward, \namecite{marvin-linzen-2018-targeted} published a new dataset in English which includes not only subject-verb agreement items, but also other dependencies (e.g. anaphora, negative polarity items) and more complex constructions. Their results showed that although the behavior of various RNNs on this dataset is far from human performance, they obtain competitive results in various settings. Taking a different approach, \namecite{lakretz-etal-2019-emergence} were able to identify individual cells on a LSTM model which encode information about grammatical number and plurality in English, suggesting that the network effectively captures some morphosyntactic information from raw text.

The growing interest in this research area motivated the release of \textit{SyntaxGym}\footnote{\url{https://syntaxgym.org/}}, an online platform for targeted evaluation of language models \cite{gauthier-etal-2020-syntaxgym}, as well as datasets in different languages, such as \namecite{mueller-etal-2020-cross} (for English, French, German, Hebrew, and Russian), \namecite{perez-mayos-etal-2021-assessing} (for Spanish), or \namecite{garciapropor22} (for Galician and Portuguese), which is the one used in this work.

Unlike LSTMs, Transformer architectures \cite{vaswani2017attention} use a non recurrent neural network which learns relations between words using a self-attention mechanism, and they can be interpreted as induced-structure models \cite{henderson-2020-unstoppable}. On a comparison of LSTM and Transformer architectures, \namecite{tran-etal-2018-importance} found that the former slightly outperform Transformers on English subject-verb agreement. In this respect, the release of large Transformer-based models, such as BERT \cite{devlin-etal-2019-bertB} and its variants, gave rise to a larger interest in exploring their linguistic abilities. Given that training these models is computationally expensive, most studies explore publicly available resources \cite{goldberg2019assessing,mueller-etal-2020-cross}. Among others, \namecite{perez-mayos-etal-2021-much} have shown that more training data yields better performance in most syntactic tasks in English, and \namecite{perez-mayos-etal-2021-assessing} compared multilingual and monolingual models in English and Spanish: the results seem to indicate that the syntactic generalization of each model type is language-specific, as some multilingual architectures work better than the monolingual ones in some scenarios and vice-versa. More recently, \namecite{garciapropor22} evaluated a variety of BERT models for Galician and Portuguese and found that monolingual ones seem to properly identify some agreement dependencies across intervening material such as relative clauses but struggle when dealing with others, like inflected infinitives.

More related to our project, \namecite{wei-etal-2021-frequency} trained BERT models for English controlling the training data, and found that word frequency during the learning phase influences the prediction performance of a model on subject-verb agreement dependencies (something which was previously suggested by \namecite{marvin-linzen-2018-targeted}). However, to the best of our knowledge, there are no studies analyzing the impact of training time on the syntactic abilities of Transformer models, possibly because intermediate training checkpoints are not often available (although \namecite{multiberts} just released several checkpoints at different training steps of BERT models for English). 

This paper presents a detailed comparison between monolingual and multilingual BERT models for Galician. On the one hand, we explore the models' behaviour regarding linguistic properties of the test items, such as the length of the target dependency or the presence of attractors. On the other hand, we compare the models' performance taking into account several parameters, such as the number of hidden layers, the amount of training data, their initialization, or the number of their training steps.

\section{Galician BERT models}
\label{sec:models}

In our experiments we compare the following monolingual and multilingual models:
\begin{itemize}
    \item \textbf{mBERT}, which is the official multilingual BERT (base, with 12 layers).
    \item \textbf{Bertinho-small} (with 6 hidden layers) and \textbf{Bertinho-base} (with 12 hidden layers) published by \namecite{PLN6319}. These two models have a vocabulary of 30k tokens, and have been trained on the Galician Wikipedia (with about 45M words).
    \item \textbf{BERT-small} (6 hidden layers) and \textbf{BERT-base }(12 layers) released by \namecite{garcia-2021-exploring}, both trained on a corpus of about 550M tokens. BERT-small has a vocabulary size of 52k tokens and has been trained during 1M steps. BERT-base has been initialized from mBERT (which includes Galician as on of its 102 languages). It has a vocabulary size of 119,547 and has been trained during 600 steps.
\end{itemize}

Additionally, we release the checkpoints of the latter two monolingual BERT models (BERT-small and BERT-base) mentioned above \cite{garcia-2021-exploring}. These models have been trained on a corpus which combines the Galician Wikipedia (April 2020 dump), SLI GalWeb \cite[composed by crawled texts from various domains]{agerri-etal-2018-developing}, CC-100 \cite{wenzek-etal-2020-ccnet}, and other data crawled from online newspapers. It was semi-automatically cleaned by removing duplicate sentences, and utterances with many symbols and punctuations. The models were trained with a masked language modeling objective on a single Titan XP GPU (12GB), with batch sizes of 208 (small) and 198 (base), and using the \textit{transformers} library \cite{wolf-etal-2020-transformers}. For each model, we saved a checkpoint every 25k steps (about 12h and 26h for the small and base models respectively) up to 425k steps.\footnote{All checkpoints are available at \url{https://github.com/marcospln/galician_bert_checkpoints}} To avoid confusion with the originally published models BERT-small and BERT-base, these newly released checkpoints will be referred to as Check-small and Check-base.

\section{Methodology}
\label{sec:method}
\subsection{Research questions and experimental design}
This work aims to explore the following research questions: 

\begin{itemize}
    \item \textbf{Q1:} Are Galician BERT LMs able to resolve agreement dependencies?
    \item \textbf{Q2:} Does model accuracy vary as a function of dependency type?
    \item \textbf{Q3:} Are Galician BERT LMs subject to interference effects from structurally-irrelevant attractor nouns?
    \item \textbf{Q4:} Does accuracy improve with training time?
\end{itemize}

\begin{table*}[!htbp]
\centering
\begin{tabular}{|l|l|l|p{9.8cm}|}
\hline
\textbf{dep}                      & \textbf{length}        & \textbf{attr} & \textbf{example}                                                                      \\ \hline \hline
\multirow{4}{*}{\textbf{noun-adj}} & \multirow{2}{*}{short} & no                 & O \textbf{neno} que xogaba onte alí é \textbf{alto}$|$*alta.                                                       \\ \cline{3-4} 
                                         &                        & yes                & O \textbf{neno} que xogaba onte alí coa \underline{nena} é \textbf{alto}$|$*alta.                                          \\ \cline{2-4} 
                                         & \multirow{2}{*}{long}  & no                 & O \textbf{neno} que xogaba no parque inaugurado recentemente é \textbf{alto}$|$*alta.                              \\ \cline{3-4} 
                                         &                        & yes                & O \textbf{neno} que xogaba no parque inaugurado recentemente coa \underline{nena} é \textbf{alto}$|$*alta.                     \\ \hline
\multirow{4}{*}{\textbf{subj-verb}}   & \multirow{2}{*}{short} & no                 & O \textbf{neno} que xogaba onte alí \textbf{aparece}$|$*aparecen na televisión.                                           \\ \cline{3-4} 
                                         &                        & yes                & O \textbf{neno} que xogaba onte alí cos outros \underline{nenos} \textbf{aparece}$|$*aparecen na televisión.                          \\ \cline{2-4} 
                                         & \multirow{2}{*}{long}  & no                 & O \textbf{neno} que xogaba no parque inaugurado recentemente \textbf{aparece}$|$*aparecen na televisión.                  \\ \cline{3-4} 
                                         &                        & yes                & O \textbf{neno} que xogaba no parque inaugurado recentemente cos outros \underline{nenos} \textbf{aparece}$|$*aparecen na televisión. \\ \hline
\end{tabular}
\caption{\label{tab:example}Sample set of the experimental conditions for noun-adjective and subject-verb agreement dependencies. \textit{dep} is the target dependency, and \textit{att} refers to the presence/absence of an attractor word. Words in bold are in a dependency relation, and underlined words are attractors, which agree with the wrong alternative marked with *. The base sentence (i.e. short without attractor) for noun-adjective dependencies means ``The boy who was playing there yesterday is tall'', and for subject-verb agreement dependencies means ``The boy who was playing there yesterday appears on TV''.}
\end{table*}

To provide an (at least tentative) answer to these questions, we created a word prediction task that we run in the different models under evaluation (i.e. mBERT, Bertinho-small and base, BERT-small and base and the different checkpoints of Check-small and base. Our task had a factorial design which manipulated the type of dependency (noun-adjective vs. subject-verb agreement), the amount of intervening material (short vs. long) and the presence or absence of an intervening but structurally irrelevant noun that mismatches in agreement features with the head word (no attractor vs attractor). A sample set of the experimental conditions is shown in Table \ref{tab:example}. 

We will pay particular attention to the models' accuracy for the experimental conditions at different steps in the training process by testing checkpoints at every 25k steps up to 425k for both Check-small and base. This will also allow us to investigate not only the effects of the training steps (on the various checkpoints) but also to make, among others, the following comparisons: (a) the impact of the training data and vocabulary size, comparing BERT-small and Bertinho-small; (b) the influence of the hidden layers, using Bertinho-base and Bertinho-small; (c) the effects of fine-tuning on monolingual data, comparing mBERT with our BERT-base (initialized from mBERT and fine-tuned in Galician). This inquiry is possible thanks to the public availability of the dataset described in the next section.

\subsection{The dataset}
In order to run the experiments described above, we have used a subset of the dataset released by \namecite{garciapropor22}\footnote{\url{https://github.com/crespoalfredo/PROPOR2022-gl-pt}} ---the first and only available dataset for targeted syntactic evaluation for Galician and Portuguese (number and gender) agreement dependencies. We limit ourselves to a subset of the dataset by choosing items with the structure of those in Table~\ref{tab:example}.

For noun-adjective agreement dependencies, the dataset contains 2,112 sentences. In order to avoid possible confounds, gender was conterbalanced so that half of the items had a feminine target and the other half a masculine one. For subject-verb agreement dependencies, the dataset contains 4,368 sentences. Similarly, number was counterbalanced so that half of the items had a singular target and the other half a plural one. It is worth noting that, overall, there are less experimental items without an attractor than those with an attractor. This is because the original dataset contained variants of the attractors to avoid potential lexical biases. Further details on the dataset building procedure are available in \namecite{garciapropor22}, but it must be noted that to create the dataset, the authors selected as target (masked) words only those forms appearing in the vocabulary of the (monolingual and multilingual) models in order to allow the evaluation of all models using the same number of experimental items. 

\subsection{Evaluation metrics}

\begin{table*}[!ht]
\begin{center}
\begin{tabular} {|l|c|c|c|c|}
  \hline
  \multicolumn{5}{|l|}{\bf Sentence}\\
  \hline
  \multicolumn{5}{|c|}{As \textbf{nenas} que xogaban onte alí co outro \underline{neno} *ten$|$\textbf{teñen} fame.}\\
  \multicolumn{5}{|c|}{\footnotesize{\it `The girls who were playing there yesterday with the other boy *is$|$are hungry.'}}\\
  \hline\hline
  {\bf Model}         & {\bf  Prob Corr} & {\bf  Prob Wrong} & {\bf 0/1 accuracy} & {\bf PD accuracy}\\ \hline
  {\it mBERT}         &            0.0007 &           0.0005  &        1  &  0.236    \\
  {\it Bertinho-base} &            0.1856 &           0.0030  &        1  &  0.968    \\
  \hline
\end{tabular}
\end{center}
\caption{\label{tab:metric}Example of a sentence where both models (mBERT and Bertinho-base) give higher probability to the correct inflection (\textit{Prob Corr} vs \textit{Prob Wrong}).}
\end{table*}

Before moving into the results of the experiments, some notes on evaluation procedures deserve mentioning. Contrary to the standard evaluation procedure which considers that a model succeeds if it assigns a higher probability to the correct than to the incorrect form (assigning thus either a 0 or a 1), \namecite{newman-etal-2021-refining} propose the use of a large set of correct/incorrect pairs on each sentence to probe a model.
In this vein, they measure the model's \textit{systematicity} (in how many pairs per sentence the model prefers the correct alternative) and \textit{likely behaviour} (the probability of generating a correct inflection).
In this method, large lists of verbs are gathered using corpus frequencies, which are then used to replace the original pairs of each context. Hence, using this strategy involves the evaluation on many potentially semantically infelicitous (or implausible) sentences.

\begin{figure}[!ht]
  \centering
  \includegraphics[width=.45\textwidth]{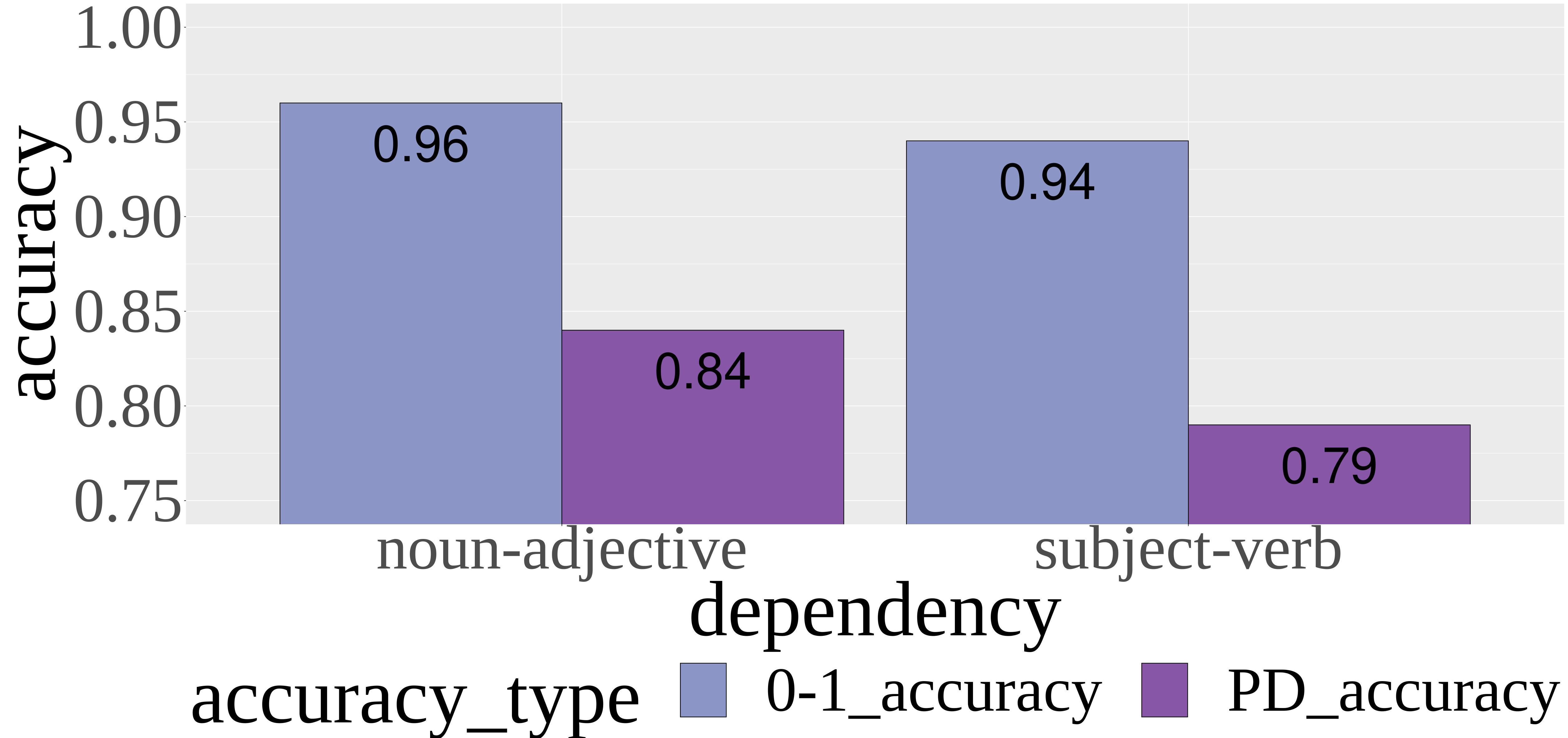}
  \caption{Mean accuracy for BERT-base's long sentences with an intervening attractor using 0-1 and PD accuracy metrics.}
  \label{fig:metric}
\end{figure}

By contrast, we put the focus on the probability distance between the correct and the incorrect alternatives provided by the dataset, and propose a new metric dubbed \textit{Probability Distance} (PD accuracy). It is calculated by normalizing the probabilities of the correct and incorrect targets (e.g. \textit{teñen} vs \textit{ten}) via percentages, and then substracting the percentage of the incorrect form from the percentage of the correct one. The rationale behind this metric is motivated by the observation that traditional binary metrics obscure possible effects because subtle differences such that between 0.51 vs 0.49 would be immediately translated to 1 or 0 (thus obtaining the same accuracy as for large differences such as 0.85 vs 0.15). Instead, PD accuracies are circumscribed to a probability space which only includes a semantically plausible target pair, and accuracy is calculated within that narrowed probability space, keeping the original distance between correct and incorrect words. To demonstrate this, Table~\ref{tab:metric} shows the results for a subject-verb long agreement dependency with an attractor for which mBERT and Bertinho-base give a higher probability for the correct inflection. Nonetheless, while 0/1 accuracy focuses on the most likely alternative, and would thus assign a 1 in both cases, PD accuracy brings the distance between the correct and incorrect alternatives to the fore. To further demonstrate this, Figure \ref{fig:metric} shows the mean accuracy for BERT-base's long sentences with an intervening attractor (i.e. examples such as the one in Table \ref{tab:metric}). We are only showing BERT-base results for long sentences with an attractor for the sake of simplicity, as these are the cases where models tend to fail. What can be observed here is that binary metrics clearly overestimate the systematicity of language models, as accuracy drops once PD accuracy is calculated. Critically, PD accuracy also provides a better threshold for comparison between models with different vocabulary sizes.

\section{Results and Discussion}
\label{sec:results}
We will first report and discuss the accuracy provided by the five available models and then, we will focus on the results for Check-small and Check-base at different training checkpoints.

\subsection{Published models}
\begin{figure}[!ht]
  \centering
  \includegraphics[width=.45\textwidth]{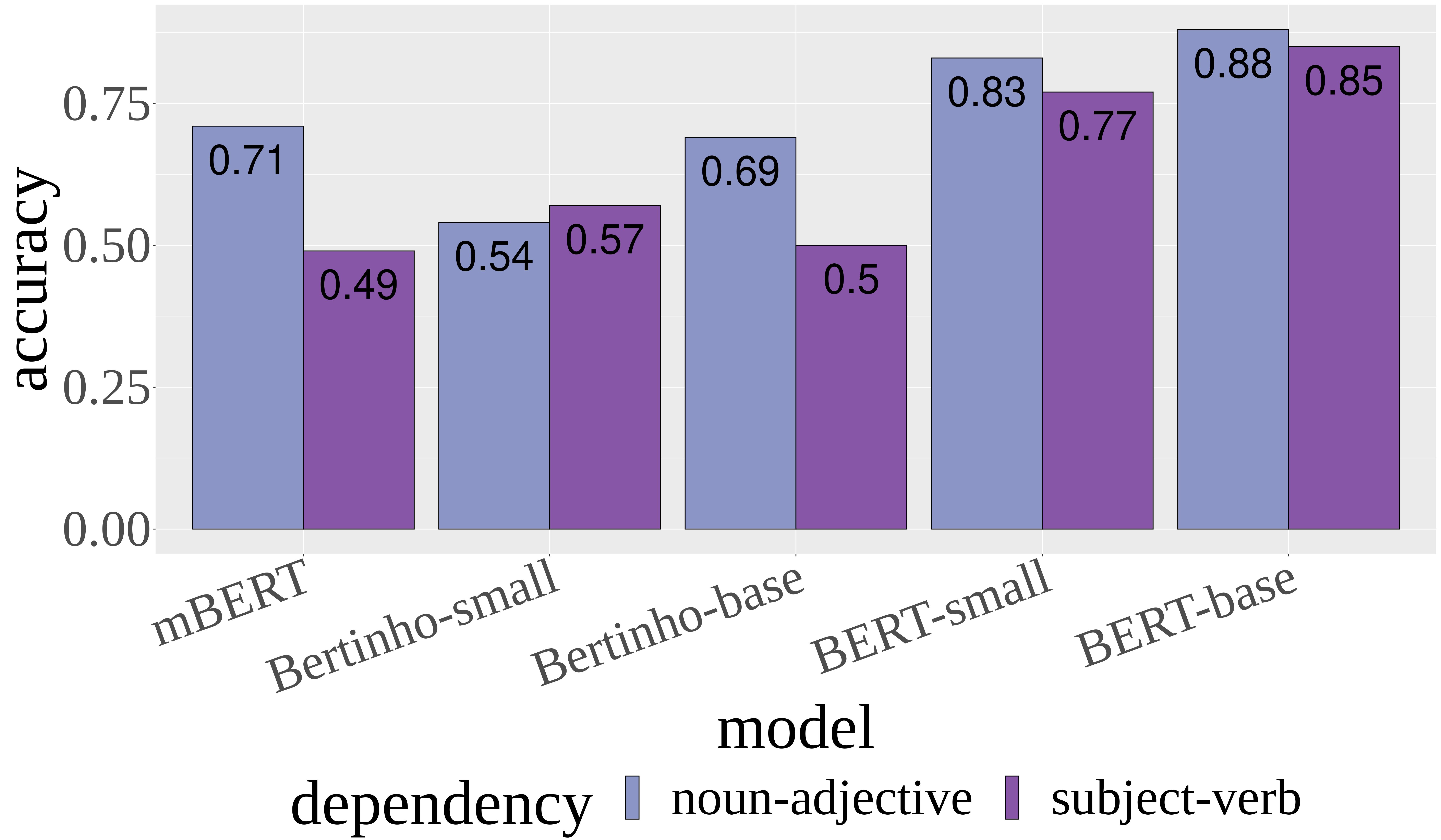}
  \caption{Mean accuracy by dependency type for the five models under investigation.}
  \label{fig:overall}
\end{figure}

\paragraph{Overall accuracy:}
Figure~\ref{fig:overall} shows the overall accuracy for each model for noun-adjective and subject-verb agreement dependencies, regardless of dependency length of the presence/absence of an attractor. Several ideas related to our research questions can be tackled at this point: first, there is a decline in accuracy for the monolingual models (BERT-base$>$BERT-small$>$Bertinho-base$>$Bertinho-small), suggesting that the amount of training data heavily influences the models' performance. Interestingly, mBERT's results resemble those from Bertinho-base, most possibly because they have been trained with the same Galician data (Wikipedia) and have a similar architecture (same number of hidden layers and dimensionality). BERT-base and BERT-small show a relatively acceptable performance, while the other three models (Bertinho-base, Bertinho-small and mBERT) are closer to chance performance (see Q1). This is particularly true for subject-verb agreement dependencies, while all models except Bertinho-small provide better results for noun-adjective dependencies (see Q2). In line with the results of \namecite{goldberg2019assessing} for English, Bertinho-small obtained slightly better accuracy than its base variant on subject-verb agreement. 

\begin{figure}[!ht]
  \centering
  \includegraphics[width=.45\textwidth]{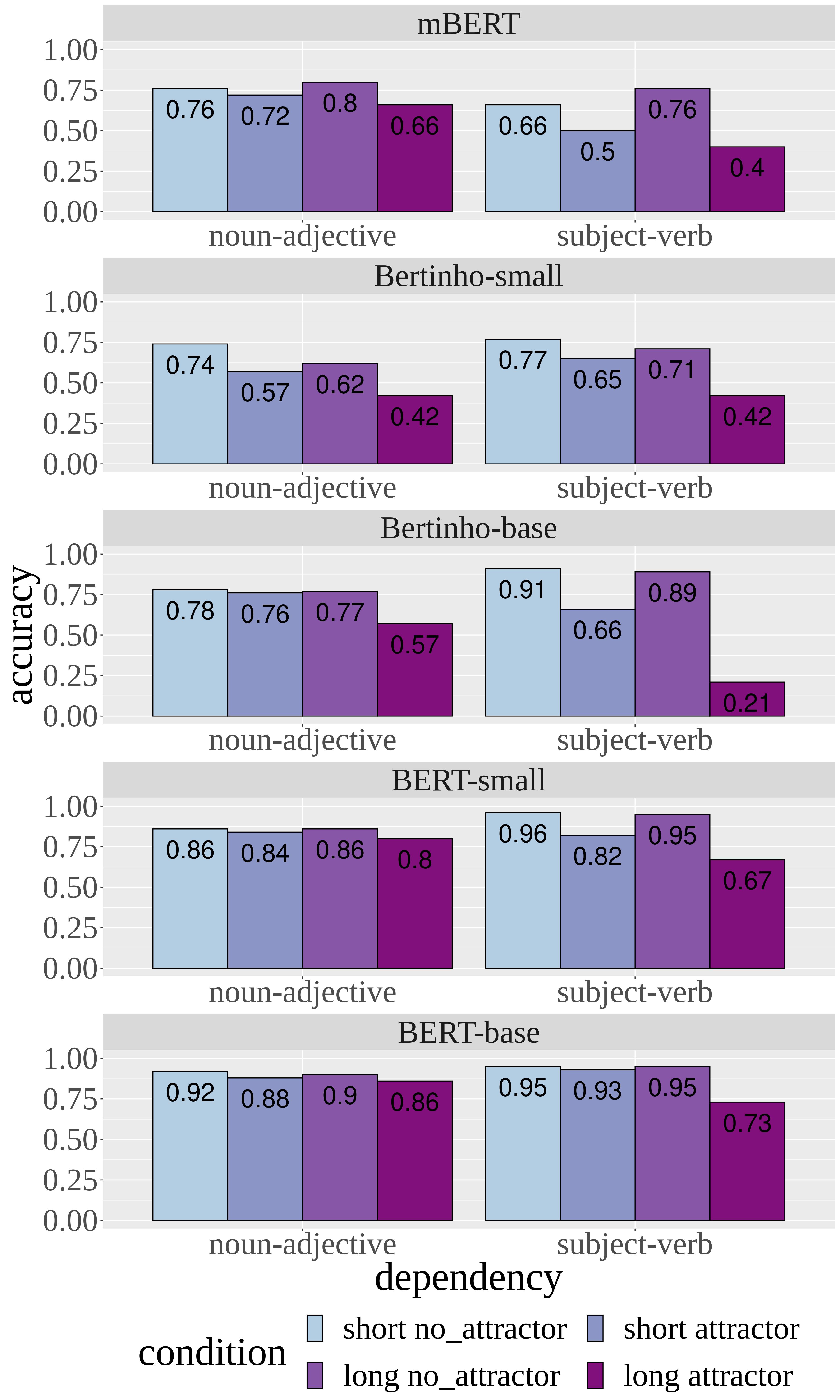}
  \caption{Mean accuracy by dependency type and experimental condition for the five models under investigation.}
  \label{fig:conditions}
\end{figure}

\begin{figure*}[!ht]
  \centering
  \includegraphics[width=.90\textwidth]{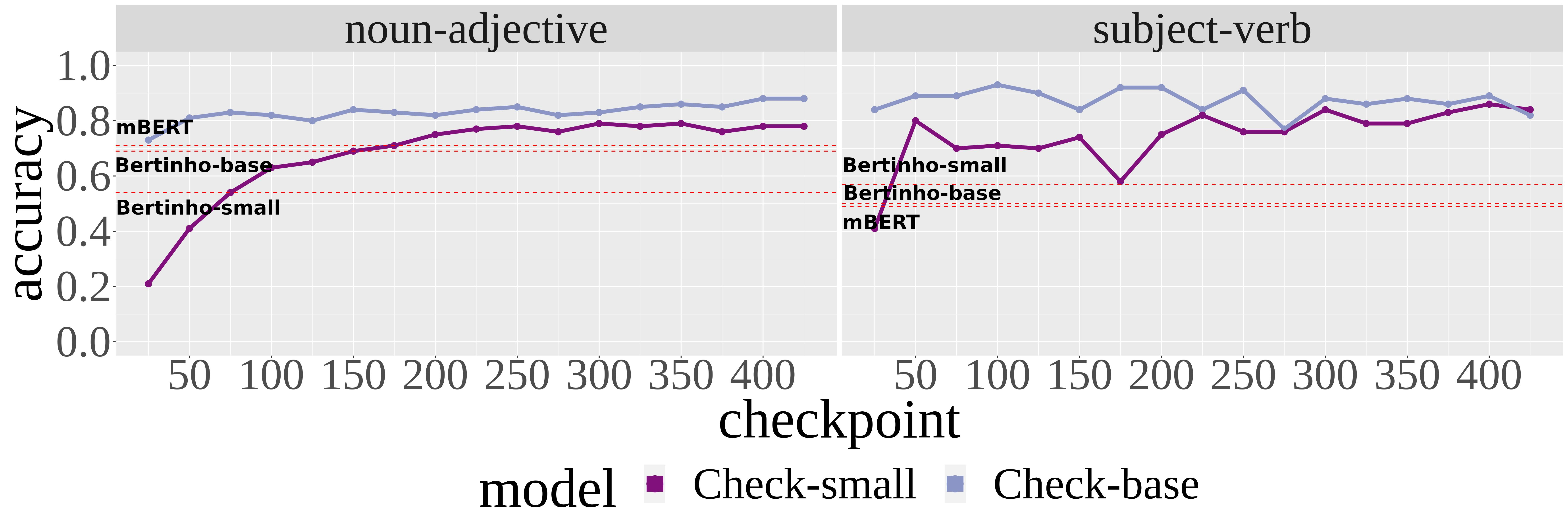}
  \caption{Mean accuracy by dependency type and checkpoint for Check-base and Check-small. The horizontal red lines indicate the overall accuracy mean for Bertinho-base, Bertinho-small and mBERT for ease of comparison (cf. Figure 2).}
  \label{fig:curve}
\end{figure*}

\paragraph{Accuracy per condition:}
Figure~\ref{fig:conditions} provides a closer picture of the models' performance for noun-adjective and subject-verb dependencies when looking at the four different experimental conditions. These results tap directly into the possible presence of agreement attraction effects (see Q3). Based on previous studies, out of the four experimental conditions, short sentences with no attractor were predicted to be the easiest ones, while long sentences with an attractor were predicted to be the hardest ones. This was borne out in the results, confirming that Galician BERT models are lured by structurally irrelevant mismatching nouns that intervene in agreement dependencies between the head and the target. Critically, the emergence of attraction effects is mediated by the distance between the head and the target such that longer dependencies are more prone to give rise to attraction effects. Nonetheless, it must be noted that not all models show equally strong attraction effects (aligning with the accuracy decline described above), and that attraction effects are steeper in subject-verb agreement dependencies than in noun-adjective agreement dependencies ---something which was foreseeable on the basis of Figure \ref{fig:overall}.

\subsection{Learning curves}

\paragraph{Overall accuracy:}
Moving now into the analyses by training checkpoints, Figure~\ref{fig:curve} shows the overall accuracy for Check-small and Check-base for the two dependencies at every checkpoint, from 25k to 425k steps.

\subparagraph{Check-small:}
Focusing on the small model (with 6 layers and trained from scratch) represented by the dark line, the results show that it needs relatively few checkpoints to surpass the average accuracy by Bertinho and mBERT. On noun-adjective dependencies, Bertinho-small is surpassed at checkpoint 75k, while it needs between 150k and 175k steps to outperform Bertinho-base and mBERT, both with 12 layers. This is even clearer on subject-verb dependencies, as the second checkpoint (75k) already shows better results than any of the three mentioned models. When comparing with the results of the published BERT-base and BERT-small (see Figure~\ref{fig:overall}), it is worth noting that these models do not obtain notoriously better results even though they have been trained for a longer period of time. These results suggest that the amount of (monolingual) training data, rather than training time, is crucial to generalize the target dependencies, as Check-small obtains better results at checkpoint 75k than Bertinho-base at 1.5M training steps.

\subparagraph{Check-base:}
Moving now into Check-base, represented by the dark line, it should be reminded that this model has been initialized with the weights of mBERT, and at the first checkpoint (25k) it already obtains comparable results to that of the final BERT-base model on subject-verb agreement (see Figure~\ref{fig:overall}). On noun-adjective dependencies, the model keeps a more constant learning rate, but it achieves similar performance than BERT-base at around 400k steps. In this case, we may hypothesize that the model is taking advantage of the linguistic properties of other languages covered by mBERT, and it adapts the model to Galician on early steps. Contrarily to the constant learning rate observed for noun-adjective dependencies, the panorama learning curve for subject-verb dependencies seems to be much more unstable. Indeed, no improvement is observed for neither Check-base nor Check-small. Although accuracy improves with time, subject-verb agreement dependencies experience ups and downs in intermediate checkpoints.

\begin{figure*}[!ht]
  \centering
  \includegraphics[width=.9\textwidth]{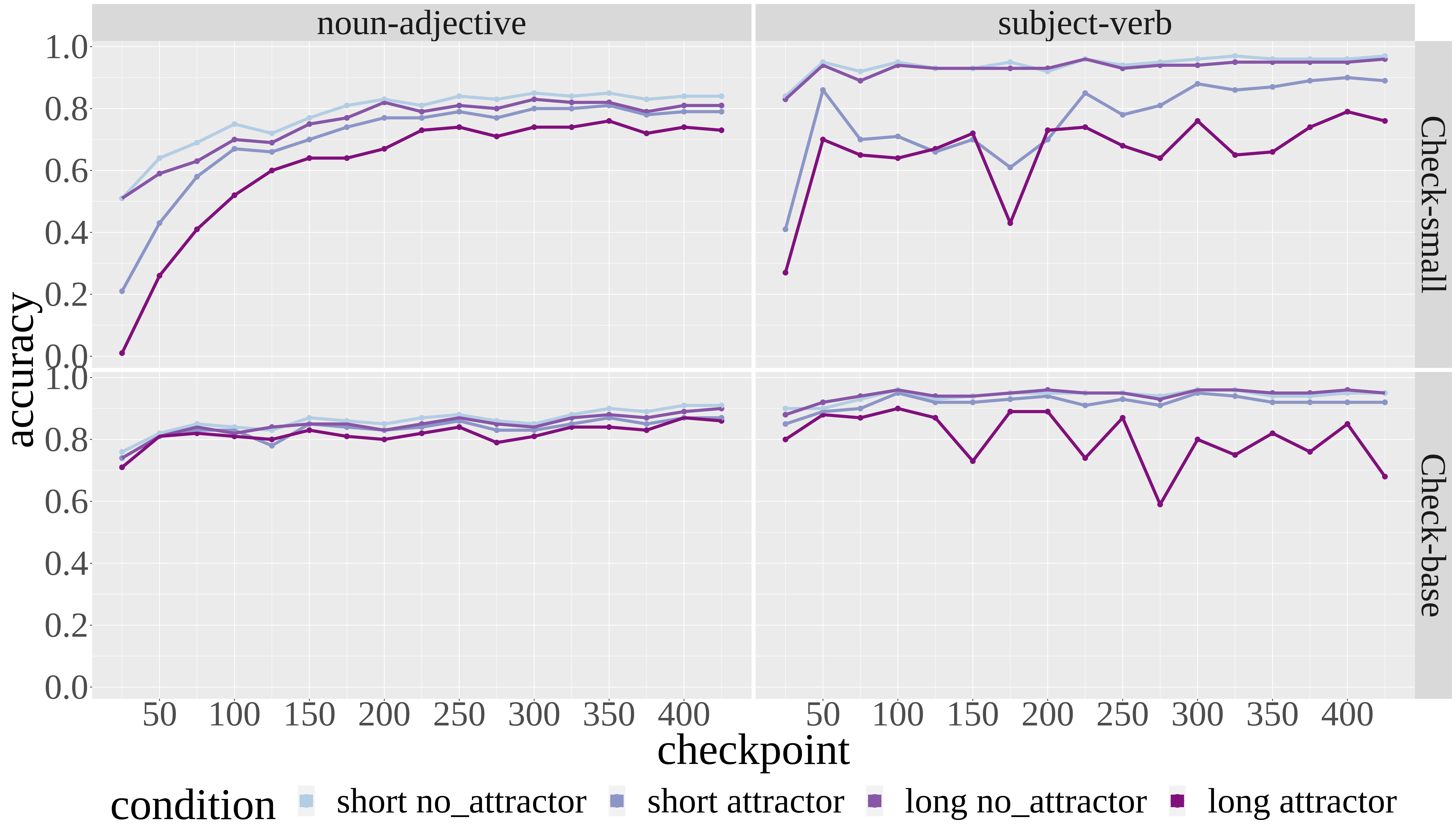}
  \caption{Mean accuracy of Check-small (top) and Check-base (bottom) by dependency type, experimental condition, and checkpoint.}
  \label{fig:curve_condition}
\end{figure*}

\paragraph{Accuracy per condition:}
Finally, Figure~\ref{fig:curve_condition} shows the curves for each experimental condition for Check-small and Check-base at the different training steps. As expected, short contexts and sentences without attractors are easily solved by both models on the two dependencies. As previously shown (Figure~\ref{fig:curve}), Check-base overtakes the performance of mBERT on the first checkpoints, but then the raise of the learning curve is very small, namely for subject-verb agreement. The small variant, especially on noun-adjective dependencies, shows a constant learning process which seems to stabilize around checkpoint 300k.
Interestingly, subject-verb agreement dependencies are easily solved by Check-small at 50k steps in the absence of attractors. However, when attractors are present, Check-small is sensitive to them even in short contexts, where it preserves the performance again about checkpoint 300k. Finally, on long-distance dependencies with attractors, none of the models seem to have a stable behaviour, as the performance of both of them varies unpredictably. This is more noticeable for Check-base, which solves most cases properly but struggles with the more complex structures (i.e. long sentences with attractors).

\section{Conclusions and future work}
\label{sec:conclusions}
This paper has presented a multidimensional evaluation of a variety of BERT models for Galician on two types of agreement dependencies, noun-adjective and subject-verb. We compared the performance of multilingual and monolingual models with diverse properties, including 6 and 12 layers variants, different sizes of training data and vocabularies, and two initializations: training from scratch, and fine-tuning a multilingual BERT on Galician data.

Our results show a gradient in the ability of Galician BERT LMs models to resolve agreement dependencies, with BERT-base being the most accurate and Bertinho-small the least accurate. Furthermore, we observed that accuracy varied as a function of dependency type, with noun-adjective agreement dependencies being easier to handle than subject-verb agreement dependencies. Interestingly, BERT LMs are subject to interference effects from structurally-irrelevant attractor nouns, and the degree of fallibility to attraction effects is inverse to accuracy (i.e. less accurate models show more attraction effects). Last and most important, although training time does seem to have a small effect on the models' accuracy, this factor is far from being comparable with the influence of the size of the training corpus.

Besides the results and analyses of the performed experiments, we contribute with new 34 checkpoints of BERT models for an understudied language, Galician, which are freely released with this paper and can hopefully contribute to foster research on languages different from English. 

This exploratory work has opened many lines of inquiry that we aim to explore in future research. On the one hand, we plan to create new datasets in Galician that do not only overcome some shortcomings observed in the one released by \namecite{garciapropor22} but also incorporate new types of linguistic relations. On the other hand, we plan to compare the results obtained for Galician with other languages in order to observe cross-linguistic differences and similarities. 

\section*{Acknowledgements}
This research was funded by the project ``Nós: Galician in the society and economy of artificial intelligence'' (Xunta de Galicia/Universidade de Santiago de Compostela), by grant ED431G2019/04 (Galician Government and ERDF), by a \textit{Ramón y Cajal} grant (RYC2019-028473-I), and by Grant ED431F 2021/01 (Galician Government).

\bibliographystyle{fullname}
\bibliography{anthology,bib}

\begin{thebibliography}{}

\bibitem[\protect\citename{Agerri \bgroup et al.\egroup
  }2018]{agerri-etal-2018-developing}
Agerri, R., X.~G{\'o}mez~Guinovart, G.~Rigau, and M.~A. Solla~Portela.
\newblock 2018.
\newblock Developing new linguistic resources and tools for the {G}alician
  language.
\newblock In {\em Proceedings of the Eleventh International Conference on
  Language Resources and Evaluation ({LREC} 2018)}, Miyazaki, Japan, May.
  European Language Resources Association (ELRA).

\bibitem[\protect\citename{Bernardy and
  Lappin}2017]{bernardy-lappin-2017-using}
Bernardy, J.-P. and S.~Lappin.
\newblock 2017.
\newblock Using deep neural networks to learn syntactic agreement.
\newblock In {\em Linguistic Issues in Language Technology, Volume 15, 2017}.
  CSLI Publications.

\bibitem[\protect\citename{Devlin \bgroup et al.\egroup
  }2019]{devlin-etal-2019-bertB}
Devlin, J., M.-W. Chang, K.~Lee, and K.~Toutanova.
\newblock 2019.
\newblock {BERT}: Pre-training of deep bidirectional transformers for language
  understanding.
\newblock In {\em Proceedings of the 2019 Conference of the North {A}merican
  Chapter of the Association for Computational Linguistics: Human Language
  Technologies, Volume 1}, pages 4171--4186, Minneapolis, Minnesota.
  Association for Computational Linguistics.

\bibitem[\protect\citename{Garcia}2021]{garcia-2021-exploring}
Garcia, M.
\newblock 2021.
\newblock Exploring the representation of word meanings in context: {A} case
  study on homonymy and synonymy.
\newblock In {\em Proceedings of the 59th Annual Meeting of the Association for
  Computational Linguistics and the 11th International Joint Conference on
  Natural Language Processing (Volume 1: Long Papers)}, pages 3625--3640,
  Online, August. Association for Computational Linguistics.

\bibitem[\protect\citename{Garcia and Crespo-Otero}2022]{garciapropor22}
Garcia, M. and A.~Crespo-Otero.
\newblock 2022.
\newblock {A Targeted Assessment of the Syntactic Abilities of Transformer
  Models for Galician-Portuguese}.
\newblock In {\em International Conference on Computational Processing of the
  Portuguese Language (PROPOR 2022)}, pages 46--56. Springer.

\bibitem[\protect\citename{Gauthier \bgroup et al.\egroup
  }2020]{gauthier-etal-2020-syntaxgym}
Gauthier, J., J.~Hu, E.~Wilcox, P.~Qian, and R.~Levy.
\newblock 2020.
\newblock {S}yntax{G}ym: An online platform for targeted evaluation of language
  models.
\newblock In {\em Proceedings of the 58th Annual Meeting of the Association for
  Computational Linguistics: System Demonstrations}, pages 70--76, Online,
  July. Association for Computational Linguistics.

\bibitem[\protect\citename{Goldberg}2019]{goldberg2019assessing}
Goldberg, Y.
\newblock 2019.
\newblock {Assessing BERT's Syntactic Abilities}.
\newblock arXiv preprint arXiv:1901.05287.

\bibitem[\protect\citename{Gulordava \bgroup et al.\egroup
  }2018]{gulordava-etal-2018-colorless}
Gulordava, K., P.~Bojanowski, E.~Grave, T.~Linzen, and M.~Baroni.
\newblock 2018.
\newblock Colorless green recurrent networks dream hierarchically.
\newblock In {\em Proceedings of the 2018 Conference of the North {A}merican
  Chapter of the Association for Computational Linguistics: Human Language
  Technologies, Volume 1 (Long Papers)}, pages 1195--1205, New Orleans,
  Louisiana, June. Association for Computational Linguistics.

\bibitem[\protect\citename{Henderson}2020]{henderson-2020-unstoppable}
Henderson, J.
\newblock 2020.
\newblock The unstoppable rise of computational linguistics in deep learning.
\newblock In {\em Proceedings of the 58th Annual Meeting of the Association for
  Computational Linguistics}, pages 6294--6306, Online, July. Association for
  Computational Linguistics.

\bibitem[\protect\citename{Hewitt and
  Manning}2019]{hewitt-manning-2019-structural}
Hewitt, J. and C.~D. Manning.
\newblock 2019.
\newblock {A} structural probe for finding syntax in word representations.
\newblock In {\em Proceedings of the 2019 Conference of the North {A}merican
  Chapter of the Association for Computational Linguistics: Human Language
  Technologies, Volume 1 (Long and Short Papers)}, pages 4129--4138,
  Minneapolis, Minnesota, June. Association for Computational Linguistics.

\bibitem[\protect\citename{Kuncoro \bgroup et al.\egroup
  }2018a]{kuncoro2018perils}
Kuncoro, A., C.~Dyer, J.~Hale, and P.~Blunsom.
\newblock 2018a.
\newblock The perils of natural behaviour tests for unnatural models: the case
  of number agreement.
\newblock {\em Learning Language in Humans and in Machines}, 5(6).
\newblock https://osf.io/9usyt/.

\bibitem[\protect\citename{Kuncoro \bgroup et al.\egroup
  }2018b]{kuncoro-etal-2018-lstms}
Kuncoro, A., C.~Dyer, J.~Hale, D.~Yogatama, S.~Clark, and P.~Blunsom.
\newblock 2018b.
\newblock {LSTM}s can learn syntax-sensitive dependencies well, but modeling
  structure makes them better.
\newblock In {\em Proceedings of the 56th Annual Meeting of the Association for
  Computational Linguistics (Volume 1: Long Papers)}, pages 1426--1436,
  Melbourne, Australia, July. Association for Computational Linguistics.

\bibitem[\protect\citename{Lakretz \bgroup et al.\egroup
  }2019]{lakretz-etal-2019-emergence}
Lakretz, Y., G.~Kruszewski, T.~Desbordes, D.~Hupkes, S.~Dehaene, and M.~Baroni.
\newblock 2019.
\newblock The emergence of number and syntax units in {LSTM} language models.
\newblock In {\em Proceedings of the 2019 Conference of the North {A}merican
  Chapter of the Association for Computational Linguistics: Human Language
  Technologies, Volume 1 (Long and Short Papers)}, pages 11--20, Minneapolis,
  Minnesota, June. Association for Computational Linguistics.

\bibitem[\protect\citename{Lin, Tan, and Frank}2019]{lin-etal-2019-open}
Lin, Y., Y.~C. Tan, and R.~Frank.
\newblock 2019.
\newblock Open sesame: Getting inside {BERT}{'}s linguistic knowledge.
\newblock In {\em Proceedings of the 2019 ACL Workshop BlackboxNLP: Analyzing
  and Interpreting Neural Networks for NLP}, pages 241--253, Florence, Italy,
  August. Association for Computational Linguistics.

\bibitem[\protect\citename{Linzen, Dupoux, and
  Goldberg}2016]{linzen-etal-2016-assessing}
Linzen, T., E.~Dupoux, and Y.~Goldberg.
\newblock 2016.
\newblock Assessing the ability of {LSTM}s to learn syntax-sensitive
  dependencies.
\newblock {\em Transactions of the Association for Computational Linguistics},
  4:521--535.

\bibitem[\protect\citename{Linzen and Leonard}2018]{linzen-brian18}
Linzen, T. and B.~Leonard.
\newblock 2018.
\newblock Distinct patterns of syntactic agreement errors in recurrent networks
  and humans.
\newblock In {\em {Proceedings of the 40th Annual Conference of the Cognitive
  Science Society}}.
\newblock arXiv preprint arXiv:1807.06882.

\bibitem[\protect\citename{Marvin and Linzen}2018]{marvin-linzen-2018-targeted}
Marvin, R. and T.~Linzen.
\newblock 2018.
\newblock Targeted syntactic evaluation of language models.
\newblock In {\em Proceedings of the 2018 Conference on Empirical Methods in
  Natural Language Processing}, pages 1192--1202, Brussels, Belgium,
  October-November. Association for Computational Linguistics.

\bibitem[\protect\citename{Mueller \bgroup et al.\egroup
  }2020]{mueller-etal-2020-cross}
Mueller, A., G.~Nicolai, P.~Petrou-Zeniou, N.~Talmina, and T.~Linzen.
\newblock 2020.
\newblock Cross-linguistic syntactic evaluation of word prediction models.
\newblock In {\em Proceedings of the 58th Annual Meeting of the Association for
  Computational Linguistics}, pages 5523--5539, Online, July. Association for
  Computational Linguistics.

\bibitem[\protect\citename{Newman \bgroup et al.\egroup
  }2021]{newman-etal-2021-refining}
Newman, B., K.-S. Ang, J.~Gong, and J.~Hewitt.
\newblock 2021.
\newblock Refining targeted syntactic evaluation of language models.
\newblock In {\em Proceedings of the 2021 Conference of the North American
  Chapter of the Association for Computational Linguistics: Human Language
  Technologies}, pages 3710--3723, Online, June. Association for Computational
  Linguistics.

\bibitem[\protect\citename{P{\'e}rez-Mayos, Ballesteros, and
  Wanner}2021]{perez-mayos-etal-2021-much}
P{\'e}rez-Mayos, L., M.~Ballesteros, and L.~Wanner.
\newblock 2021.
\newblock How much pretraining data do language models need to learn syntax?
\newblock In {\em Proceedings of the 2021 Conference on Empirical Methods in
  Natural Language Processing}, pages 1571--1582, Online and Punta Cana,
  Dominican Republic, November. Association for Computational Linguistics.

\bibitem[\protect\citename{P{\'e}rez-Mayos \bgroup et al.\egroup
  }2021]{perez-mayos-etal-2021-assessing}
P{\'e}rez-Mayos, L., A.~T{\'a}boas~Garc{\'\i}a, S.~Mille, and L.~Wanner.
\newblock 2021.
\newblock Assessing the syntactic capabilities of transformer-based
  multilingual language models.
\newblock In {\em Findings of the Association for Computational Linguistics:
  ACL-IJCNLP 2021}, pages 3799--3812, Online, August. Association for
  Computational Linguistics.

\bibitem[\protect\citename{Sellam \bgroup et al.\egroup }2022]{multiberts}
Sellam, T., S.~Yadlowsky, J.~Wei, N.~Saphra, A.~D'Amour, T.~Linzen,
  J.~Bastings, I.~Turc, J.~Eisenstein, D.~Das, I.~Tenney, and E.~Pavlick.
\newblock 2022.
\newblock {The MultiBERTs: BERT Reproductions for Robustness Analysis}.
\newblock In {\em {The Tenth International Conference on Learning
  Representations (ICLR 2022)}}.
\newblock arXiv preprint arXiv:2106.16163.

\bibitem[\protect\citename{Tran, Bisazza, and
  Monz}2018]{tran-etal-2018-importance}
Tran, K., A.~Bisazza, and C.~Monz.
\newblock 2018.
\newblock The importance of being recurrent for modeling hierarchical
  structure.
\newblock In {\em Proceedings of the 2018 Conference on Empirical Methods in
  Natural Language Processing}, pages 4731--4736, Brussels, Belgium,
  October-November. Association for Computational Linguistics.

\bibitem[\protect\citename{Vaswani \bgroup et al.\egroup
  }2017]{vaswani2017attention}
Vaswani, A., N.~Shazeer, N.~Parmar, J.~Uszkoreit, L.~Jones, A.~N. Gomez,
  L.~Kaiser, and I.~Polosukhin.
\newblock 2017.
\newblock {Attention Is All You Need}.
\newblock arXiv preprint arXiv:1706.03762.

\bibitem[\protect\citename{Vilares, Garcia, and
  Gómez-Rodríguez}2021]{PLN6319}
Vilares, D., M.~Garcia, and C.~Gómez-Rodríguez.
\newblock 2021.
\newblock {Bertinho: Galician BERT Representations}.
\newblock {\em Procesamiento del Lenguaje Natural}, 66:13--26.

\bibitem[\protect\citename{Wei \bgroup et al.\egroup
  }2021]{wei-etal-2021-frequency}
Wei, J., D.~Garrette, T.~Linzen, and E.~Pavlick.
\newblock 2021.
\newblock Frequency effects on syntactic rule learning in transformers.
\newblock In {\em Proceedings of the 2021 Conference on Empirical Methods in
  Natural Language Processing}, pages 932--948, Online and Punta Cana,
  Dominican Republic, November. Association for Computational Linguistics.

\bibitem[\protect\citename{Wenzek \bgroup et al.\egroup
  }2020]{wenzek-etal-2020-ccnet}
Wenzek, G., M.-A. Lachaux, A.~Conneau, V.~Chaudhary, F.~Guzm{\'a}n, A.~Joulin,
  and E.~Grave.
\newblock 2020.
\newblock {CCN}et: Extracting high quality monolingual datasets from web crawl
  data.
\newblock In {\em Proceedings of the 12th Language Resources and Evaluation
  Conference}, pages 4003--4012, Marseille, France, May. European Language
  Resources Association.

\bibitem[\protect\citename{Wolf \bgroup et al.\egroup
  }2020]{wolf-etal-2020-transformers}
Wolf, T., L.~Debut, V.~Sanh, J.~Chaumond, C.~Delangue, A.~Moi, P.~Cistac,
  T.~Rault, R.~Louf, M.~Funtowicz, J.~Davison, S.~Shleifer, P.~von Platen,
  C.~Ma, Y.~Jernite, J.~Plu, C.~Xu, T.~Le~Scao, S.~Gugger, M.~Drame, Q.~Lhoest,
  and A.~Rush.
\newblock 2020.
\newblock Transformers: State-of-the-art natural language processing.
\newblock In {\em Proceedings of the 2020 Conference on Empirical Methods in
  Natural Language Processing: System Demonstrations}, pages 38--45, Online,
  October. Association for Computational Linguistics.

\end{thebibliography}

\end{document}